\documentclass{article}
\usepackage{spconf,amsmath,graphicx}
\usepackage{graphicx}
\usepackage{caption}
\usepackage{booktabs}
\usepackage{multirow}
\usepackage{bigstrut}
\usepackage{amsmath}
\usepackage{amstext}
\usepackage{hyperref}

\title{Unsupervised Multiple Choices Question Answering \\ via Universal Corpus}

\name{Qin Zhang$^1$\thanks{
This research was supported by National Natural Science Foundation of China (62206179, 92270122),
Guangdong Provincial Natural Science Foundation (2022A1515010129, 2023A1515012584), 
University stability support program of Shenzhen (20220811121315001), 
Shenzhen Research Foundation for Basic Research, China (JCYJ20210324093000002). 
}, Hao Ge$^1$, Xiaojun Chen$^1$, Meng Fang$^2$
}
\address{$^1$Shenzhen University, China, $^2$University of Liverpool, United Kingdom \\ \{qinzhang, gehao2022, xjchen\}@szu.edu.cn, Meng.Fang@liverpool.ac.uk}

\begin{document}

\maketitle
\begin{abstract}
Unsupervised question answering is a promising yet challenging task, which alleviates the burden of building large-scale annotated data in a new domain. 
It motivates us to study the unsupervised multiple-choice question answering (MCQA) problem.
In this paper, we propose a novel framework designed to generate synthetic MCQA data barely based on contexts from the universal domain without relying on any form of manual annotation. Possible answers are extracted and used to produce related questions, then we leverage both named entities (NE) and knowledge graphs to discover plausible distractors to form complete synthetic samples.
Experiments on multiple MCQA datasets demonstrate the effectiveness of our method.
\end{abstract}
\begin{keywords}
Natural Language Processing, Unsupervised Multiple Choices Question Answering, Knowledge Graphs
\end{keywords}
\section{INTRODUCTION}
\label{sec:intro}

Question Answering (QA) is an important topic in natural language understanding \cite{cao-etal-2019-bag,huang-etal-2021-dagn,zhang2022survey}.
In QA, Multiple Choices Question Answering (MCQA) tasks require the model to select the answer from a set of answer candidates by employing reasoning~\cite{yu2022context, liu2023narrow, zhang2023joint,zhang2022survey,fang2014llm}.
A common approach is fine-tuning a pre-trained language model on a task-specific dataset~\cite{fabbri-etal-2020-template}. 
However, such task-specific datasets are scarce as they are only available for limited domains and languages~\cite{lewis2019unsupervised}. It means we need to derive a large number of annotated samples before applying this process to a new domain, which is time-consuming and resource-intensive \cite{fang-etal-2017-learning}.

Recently, some unsupervised methods have been proposed for Extractive Question Answering (EQA) tasks. E.g., Lewis et al \cite{lewis2019unsupervised} explored several unsupervised methods for generating question-answer pairs and showed that the obtained data could ensure satisfactory model performance, being comparable to the original data. Fabbri et al
\cite{fabbri-etal-2020-template} and Li et al \cite{li-etal-2020-harvesting} further extended this idea with template-based question generation and iterative data refinement, but are still only applicable to EQA tasks. 
There are also some trials for MCQA without supervision. 
Liu and Lee \cite{liu-lee-2021-unsupervised} assumed the absence of correct answer labels, but directly train a QA model based on the context, question, and answer candidate sets. 
Ren and Zhu \cite{ren2020knowledge} emphasized the distractor generation, trying to construct a complete sample using the given context, question, as well as the correct answer. 
Nevertheless, they still depend on a certain amount of data in the target domain, like the contexts and questions, which further limits their application scenarios.

In this paper, we propose a two-stage unsupervised MCQA framework under a special case, where no labeled sample but only a universal corpus is available. We aim to construct natural questions, correct answers, and related contexts in an unsupervised manner, further generating plausible and reliable distractors as the answer candidate set.
Motivated by the recent progress in \emph{Unsupervised Extractive Question Answering}~\cite{lewis2019unsupervised}, we generate question-answer pairs in the first stage. Named entities (NE) from the context are extracted and treated as the ``correct'' answers. Then questions will be generated in a cloze-filling way via unsupervised machine translation models, yielding a series of QA pairs.
In the second stage, we will introduce answer distractors. Here we propose a hybrid method to generate high-quality distractors with the aid of both NEs and knowledge graphs (KGs), which will be used to answer the candidate set. 
In experiments, we show that our unsupervised MCQA method can achieve good results to some extent using RoBERTa~\cite{liu2019roberta} as the backbone. We also illustrate that the quality of answer distractors matters, and our hybrid generation approach contributes to better performance.

\begin{figure}[t!]
  \centering
  \includegraphics[width=0.48\textwidth]{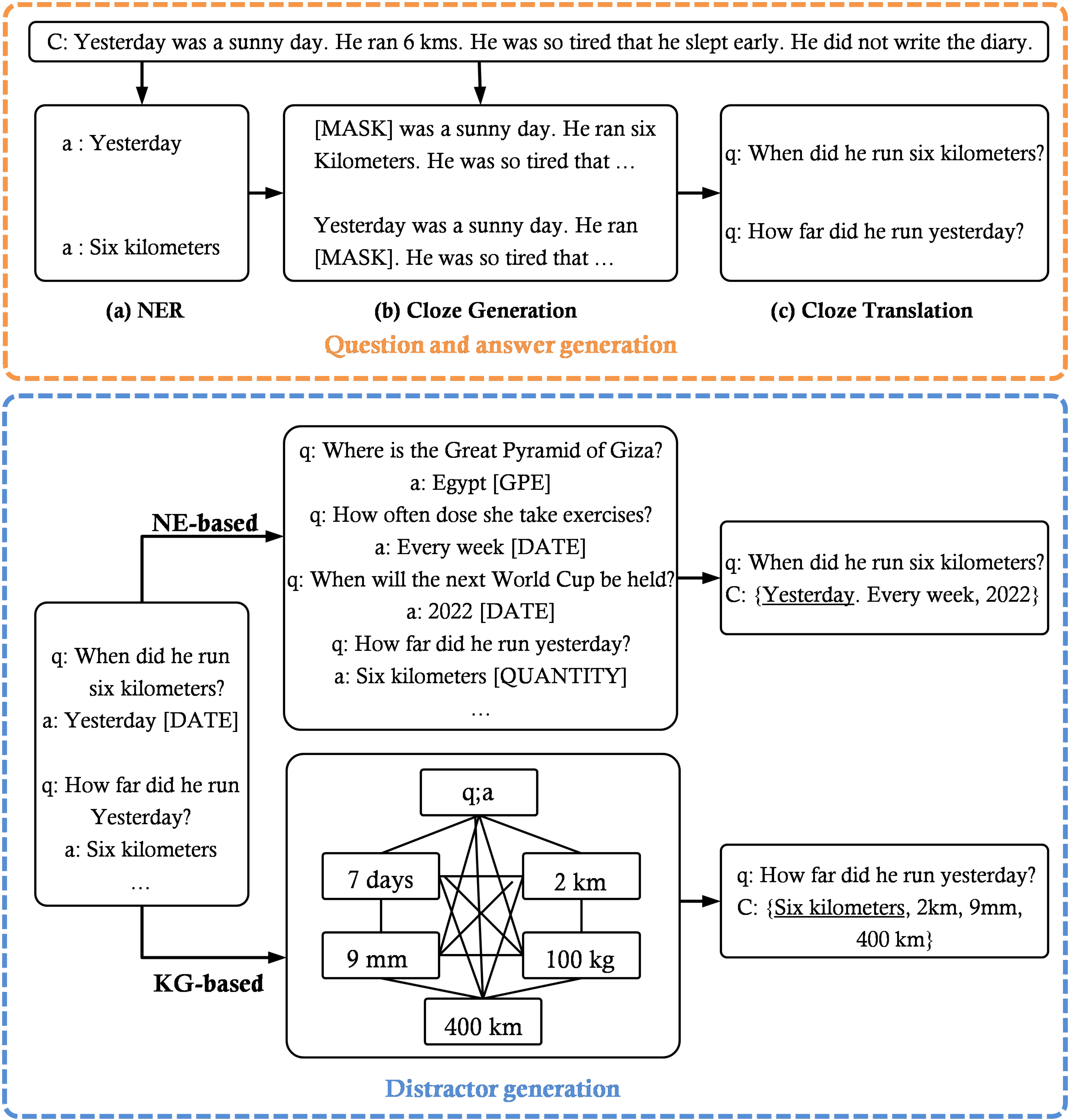}
  \caption{An overview of our method. In the first stage, we extract the answers aa from the context cc, then generate their corresponding questions qq. In the second stage, we use a hybrid method, KG-NE, to generate distractors, thus building the answer candidate set $\mathcal{C}$.
  }
  \label{img} 
\end{figure}

Our contributions are summarized as follows. Firstly, we are among the first to study the unsupervised MCQA task without data in the target domain and propose a two-stage approach equipped with QA pairs generation and distractors generation. Secondly, our extensive experiments verify the validity of our method, also the impact of different answer distractor generation methods.
\section{METHOD}
\label{sec:method}

Previous works in unsupervised MCQA~\cite{ren2020knowledge, offerijns2020better, banerjee2020self} usually assume the availability of a certain amount of data in the target domain, such as questions, answer candidates or correct answers, lowering their applicability. Our setting becomes more challenging but meaningful, where target data is given.
It is required to construct a set of MCQA samples barely using the contexts from a universal corpus. Each sample consists of the context cc, the question qq, and the set of answer candidates $\mathcal{C}$ where the correct answer aa is labeled. Figure \ref{img} shows an overview of our unsupervised MCQA method. There are two stages included: 1) we use an extractive way to generate the question qq and its corresponding answer aa; 2) we generate distractors to construct the answer candidate set $\mathcal{C}$, thus obtaining an MCQA sample.
\subsection{Question and Answer Generation}
\label{section_qa_generation}

Similar to unsupervised EQA~\cite{lewis2019unsupervised}, we start to build QA samples from a task-agnostic open-domain source corpus. We identify all the named entities (NE) with specific NER tags (in Table~\ref{appendix_ne_types}), and treat them as the correct answers for potential questions. Such an extraction process could be conducted via open-source NLP libraries (e.g.,  
spaCy~\cite{spacy}) without extra training. 
Then for each extracted answer, we generate a question to form a question-answer pair.
To this end, we first mask the NE-like answer to obtain the cloze from the context, then use a way similar to the machine translation to transform the cloze to the natural question~\cite{dou2022zero}. 
We adopt a seq2seq-based NMT model~\cite{lample2018phrase} trained on nonparallel corpora of clozes and questions to conduct such a translation task. We will generate five types of questions: \emph{who}, \emph{where}, \emph{what}, \emph{when}, and \emph{how} (e.g, ``how long'', ``how many''), based on the types of entities. 

\begin{table}[htbp]
  \centering
  \resizebox{0.46\textwidth}{!}{
    \begin{tabular}{ll}
    \toprule[1pt]
    TYPE  & DESCRIPTION \bigstrut\\
    \hline
    PERSON & People, including fictional. \bigstrut[t]\\
    NORP  & Nationalities or religious or political groups. \\
    FAC   & Buildings, airports, highways, bridges, etc. \\
    ORG   & Companies, agencies, institutions, etc. \\
    GPE   & Countries, cites, states. \\
    LOC   & Non-GPE locations, mountain ranges, bodies of water.  \\
    PRODUCT & Objects, vehicles, foods, etc. (Not services.) \\
    EVENT & Named hurricanes, battles, wars, sports events, etc. \\
    WORK\_OF\_ART & Titles of books, songs, etc. \\
    LAW   & Named documents made into laws. \\
    LANGUAGE & Any named language. \\
    DATE  & Absolute or relative dates or periods. \\
    TIME  & Times smaller than a day. \\
    PERCENT & Percentage, including (\%). \\
    MONEY & Monetary values, including unit. \\
    QUANTITY & Measurements, as of weight or distance. \bigstrut[b]\\
    \toprule[1pt]
    \end{tabular}%
}
  \caption{The type of the named entities.}    
  \label{appendix_ne_types}
\end{table}%

\subsection{Distractor Generation}
\label{section_distractor_generation}

Given a couple of a question and the corresponding answer, we seek methods to generate answer distractors as its candidate set.
A straightforward way is the \textbf{Random} method, where we randomly select the answers from other questions and treat them as distractors for the current QA pair. 
However, according to Pho et al \cite{pho2014multiple}, there should be high syntactic and semantic homogeneity between the answer and distractors to make the task more difficult. So a good distractor should have the same NE type and similar semantic meaning to the correct one. 
We thus provide several other methods with the goal of generating high-quality distractors, 
1) \textbf{NE(named entity)-based} or 2) \textbf{KG(knowledge graph)-based} methods that the generated candidates are aware of the NE type and the semantic meaning, respectively, and 3) a hybrid \textbf{NE-KG} approach combining the merits of both NE and KG. \\ \\
\textbf{NE-based.}~ A simple method selects the distractors that have the same NE type as the gold answer. Since we have already identified the NER tags during answer generation, it is realized by sampling the answers belonging to other questions as the distractors, while they share the same NE type. \\ \\
\textbf{KG-based.}~ 
One drawback of the NE-based method is that the answer candidates having the same NE type may not ensure sufficient semantic similarity, possibly lowering the challenge for QA models. Motivated by the works in former distractor generation~\cite{ren2020knowledge} and knowledge graph-based question answering~\cite{yasunaga2021qa}, we address this issue with the aid of an external knowledge base. Specifically, we use ConceptNet~\cite{speer2017conceptnet}, which is a general domain knowledge graph, as our knowledge base. We concatenate the question and the answer to build the input representation vector, then we follow Feng et al \cite{feng2020scalable} to retrieve a subgraph of the ConceptNet consisting of entities that are closely related to the input. Based on the subgraph, we use a pre-trained language model to further estimate the relevance between the entities and the input. Then entities with top-K largest scores will be regarded as the selected distractors. \\ \\
\textbf{KG-NE.}~ 
We provide a hybrid method, KG-NE, aiming at a combination of the benefits of both KG-based and NE-based methods.
The distractors generated by NE-based method may vary largely in their semantic meanings. Although the KG-based approach can provide distractors that have high semantic relevance to the answer, sometimes the KG may fail to recognize the entities in the input, making it hard to conduct the follow-up subgraph retrieval and relevance scoring operations. 
Among the five types of generated questions (Who, What, Where, When, and How), we observe that the KG-based approach only works well for the \textit{How}-type questions (refer to Table~\ref{tab:analysis2} for details).  
Therefore, we select specific generation methods for different question types in KG-NE. 
For the questions belonging to the \emph{how}-type, we apply the KG-based method, while leaving those belonging to the rest four types to the NE-based generation. %

\section{EXPERIMENTS}
\label{sec:Experiments}
\subsection{Setup} 
\textbf{Implementation.} Similar to Lewis et al \cite{lewis2019unsupervised}, we use data from the English Wikipedia for constructing the synthetic datasets. We use Spacy for NER in answer extraction and adopt the unsupervised NMT model provided by Lewis et al \cite{lewis2019unsupervised} for the question generation. Regarding the distractor generation, we adopt the ConceptNet and the RoBERTa-large model~\cite{liu2019roberta} provided by Yasunaga et al \cite{yasunaga2021qa} for constructing the entity graph and obtaining the relevance score, respectively. The synthetic datasets are derived from 101,500 passages, where we use 100,000 and 1,500 passages as the training and development set, respectively. 

We use the RoBERTa-base model \cite{liu2019roberta} as the backbone of QA model in our evaluation. 
For each of the synthetic / annotated datasets, we train a model for 3 epochs with a batch size of 16. We use the Adam optimizer~\cite{kingma2014adam} with a learning rate of 5e-5. 

We fine-tuned the Llama 2-7B~\cite{touvron2023llama2} using 50,000 samples generated by the KG-NE method with LoRA~\cite{hu2021lora}. The fine-tuning process lasted for 3 epochs, with a batch size of 4 and a learning rate of 1e-4.

\noindent \textbf{Evaluation datasets.} We use four annotated MCQA datasets, \textbf{SWAG}~\cite{zellers2018swag}, \textbf{ARC}~\cite{Clark2018ThinkYH}, \textbf{CommensenseQA}~\cite{talmor2018commonsenseqa}, and \textbf{SocialIQA}~\cite{sap2019socialiqa}. We denote all of them as ``annotated datasets'' for clarity. \footnote{The testing sets for SWAG and CommensenseQA are half of the validation sets.}  

\noindent \textbf{Baselines.} Besides the proposed methods, we consider another three baselines: ``Random'', which is the simple random method mentioned in Section~\ref{section_distractor_generation}; Sliding Window (SW) that calculates the overlap between options and questions and contexts to obtain the answer \cite{richardson2013mctest}; Knowledge Representation Learning (KRL) a zero-shot method proposed by Banerjee and Baral et al \cite{banerjee2020self}. 
For large language models (LLMs), we considered Llama-7B~\cite{touvron2023llama}, Llama 2-7B, Llama 2-13B, and ChatGPT-3.5-Turbo.

\begin{table}[htp]
    \centering
  \resizebox{0.49\textwidth}{!}{
    \setlength{\tabcolsep}{2pt}
    \begin{tabular}{ccccc}
    \toprule[1pt]
    Method & ARC-Easy & SWAG & CommensenseQA & SocialIQA \\
    \hline
    Random   & 37.2 \scalebox{0.8}{$\pm$} 1.3 & 37.0 \scalebox{0.8}{$\pm$} 2.4 & 42.2 \scalebox{0.8}{$\pm$} 1.6 & 36.2 \scalebox{0.8}{$\pm$} 1.1 \\
    NE-based & 38.3 \scalebox{0.8}{$\pm$} 0.8 & 46.0 \scalebox{0.8}{$\pm$} 2.0 & 39.6 \scalebox{0.8}{$\pm$} 1.0 & 38.9 \scalebox{0.8}{$\pm$} 0.8 \\
    KG-based & 31.0 \scalebox{0.8}{$\pm$} 0.6 & 41.4 \scalebox{0.8}{$\pm$} 2.7 & 26.8 \scalebox{0.8}{$\pm$} 1.1 & 37.2 \scalebox{0.8}{$\pm$} 0.4 \\
    KG-NE   & \textbf{38.6 \scalebox{0.8}{$\pm$} 0.4} & \textbf{49.9 \scalebox{0.8}{$\pm$} 0.6} & \textbf{42.8 \scalebox{0.8}{$\pm$} 1.1} & \textbf{39.5 \scalebox{0.8}{$\pm$} 0.5}  \\
    \toprule[1pt]
    \end{tabular}
  }
    \caption{
    The accuracy (\%) of QA models on the four benchmark test sets
    after training on synthetic datasets generated by different approaches. 
    The results are averaged over 3 random seeds along with the standard deviations. 
    }
    \label{tab:main1}
\end{table}

\begin{table}[htp]
  \centering
  \resizebox{0.49\textwidth}{!}{
  \setlength{\tabcolsep}{2pt}
    \begin{tabular}{ccccc}
    \toprule[1pt]
    Method & ARC-Easy & SWAG & {\small CommensenseQA} & SocialIQA \bigstrut[t]\\
    \hline
    Random   & 38.2 \scalebox{0.8}{$\pm$} 1.4 & 45.5 \scalebox{0.8}{$\pm$} 5.1 & 42.5 \scalebox{0.8}{$\pm$} 1.9 & 35.2 \scalebox{0.8}{$\pm$} 0.8 \bigstrut[t]\\
    SW \cite{richardson2013mctest} & 24.8 \scalebox{0.8}{$\pm$} 5.8 &  38.0 \scalebox{0.8}{$\pm$} 2.1 & 22.4 \scalebox{0.8}{$\pm$} 4.6 & 32.8 \scalebox{0.8}{$\pm$} 1.4 \\
    KRL \cite{banerjee2020self} & 33.0 & - & 38.8 & \textbf{ 48.5 } \\
    NE-based & \textbf{40.8 \scalebox{0.8}{$\pm$} 1.0} & 48.5 \scalebox{0.8}{$\pm$} 0.7 & 39.1 \scalebox{0.8}{$\pm$} 0.7 & 42.4 \scalebox{0.8}{$\pm$} 0.9 \\
    KG-based & 30.7 \scalebox{0.8}{$\pm$} 1.3 & 43.1 \scalebox{0.8}{$\pm$} 2.0 & 27.4 \scalebox{0.8}{$\pm$} 1.1 & 37.8 \scalebox{0.8}{$\pm$} 0.4 \\
    KG-NE   & 39.5 \scalebox{0.8}{$\pm$} 0.2 & \textbf{53.3 \scalebox{0.8}{$\pm$} 0.5} & \textbf{43.7 \scalebox{0.8}{$\pm$} 1.4} & 41.6 \scalebox{0.8}{$\pm$} 0.2  \\
    Llama 2-7B* & 77.7 \scalebox{0.8}{$\pm$} 0.6 & 55.8 \scalebox{0.8}{$\pm$} 1.5 & 58.2 \scalebox{0.8}{$\pm$} 0.5 & 53.6 \scalebox{0.8}{$\pm$} 0.9 \\
    \hline
    \hline
    Llama-7B & 28.34 & 27.23 & 22.68 & 25.71 \\
    Llama 2-7B & 32.46 & 25.41 & 25.37 &31.22 \\
    Llama 2-13B & 72.31 & 41.66 & 40.95 & 48.30 \\
    ChatGPT & 84.8 & 75.3 & 73.1 & 71.6 \\
    \hline
    \hline
    Supervised & 94.7$^\dagger$ & 94.1$^\dagger$ & 83.3$^\dagger$ & 84.3$^\dagger$ \\
    \toprule[1pt]
    \end{tabular}
    }
    \caption{
    The accuracy(\%) of QA models (RoBERTa-base and Llama 2-7B) after training on different synthetic datasets, where models are determined by the validation performance on the original \textsl{annotated} datasets. The results are averaged over 3 random seeds. Llama 2-7B* is model obtained by fine-tuning the Llama 2-7B model using the synthetic data (KG-NE) generated. Llama-7B, Llama 2-7B, Llama 2-13B, and ChatGPT show the results obtained directly from the original LLMs.  $\dagger$: The results are from several leaderboards, including \href{https://leaderboard.allenai.org/arc_easy/submissions/public}{Leaderboard-ARC-E}, \href{https://leaderboard.allenai.org/swag/submissions/public}{-SWAG}, \href{https://www.tau-nlp.org/csqa-leaderboard2}{-CSQA} and \href{https://leaderboard.allenai.org/socialiqa/submissions/public}{-SocialIQA}
    }
    \label{tab:main2}
\end{table}
\subsection{Results}

Table~\ref{tab:main1} shows the QA models' performance after training on the synthetic datasets generated by different methods.
Note that we use synthetic development to determine the model for the evaluation, as we assume that no sample in the target domain is available. 
Other than over-fitting, such performance gap could be attributed to the large domain gap between the source dataset (English Wikipedia) and the target datasets (\textbf{SWAG}, \textbf{ARC}, \textbf{CommonsenseQA} and \textbf{SocialIQA}). 
Those with consideration of the types of NE during the distractor generation (``NE-based'', ``KG-NE''), achieve better testing performance than other approaches. 
In particular, the hybrid method, ``KG-NE'', further improves ``NE-based'' by margin, showing the effectiveness of considering both the NE-types and the semantic meanings. 
However, solely adopting the KG-based method leads to a performance drop. According to Section \ref{section_distractor_generation}, sometimes the KG-based method may fail to generate the MCQA data from the passage, resulting in a much smaller training set. Besides, we may further investigate the cause of the poor performance of the KG-based method from the perspective of generated question types, please refer to Section \ref{section_question_types} for details. 
We also observe that the SW method, although effective in its original setting (only the correct answer $a$ is not provided), performs badly in our setting, where the model has to construct the MCQA data from scratch to train itself. Under such a challenging setting, the proposed KG-NE method shows robustness and yields the highest performance. 
 
\subsection{Additional Analysis}
\textbf{Using annotated dev sets.}~
Table \ref{tab:main2} shows the results when we slightly alleviate the strict data availability and use the \textsl{development sets from annotated data} for model selection. 
In general, the NE-based method benefits from this setting, which shows promotions to the results in Table \ref{tab:main1}. On the other hand, the KG-based method encounters a performance drop, and further partly affects the performance of the hybrid method. 
Compared to the SW method, all three of our approaches outperform it. Compared to the KRL method, we perform worse on SocialIQA, but we perform better on ARC-Easy and SWAG. Overall, our methods are superior to both baselines. However, We also compare our unsupervised methods with the state-of-the-art supervised models, indicating that there's still a large improvement that could be made, which we leave as a future direction. \\ \\
\textbf{Large language models.}~ For LLMs, in Table \ref{tab:main2} we can see the proposed corpus can improve LLM's performance significantly. When directly testing on original LLMs, they failed a lot due to the hallucinatory and struggled to understand the information behind the questions.  As a result, their performance (Llama-7B, Llama 2-7B) was even worse than the results of the fine-tuned RoBERTa-base. Surprisingly, after fine-tuning Llama 2-7B with the corpus we generated, it achieved impressive results, even surpassing Llama 2-13B on all four datasets. \\ \\
\textbf{Generated question types.}~
\label{section_question_types}
In Table \ref{tab:analysis2},
it shows that the KG-based method results in severely unbalanced in question types. It is prone to generating \textit{How}-type questions, while only a small portion of data samples belonging to the \textit{Who}-type, the \textit{Who}-type and the \textit{What}-type are produced. Such an unbalanced distribution may be another reason for the poor performance of QA models in this case. Instead, the hybrid method is conducive to more balanced question types, benefiting QA models trained on samples derived from it. \\ \\
\begin{table}[t]
\setlength{\belowcaptionskip}{-0.1cm}
  \centering
  \resizebox{0.38\textwidth}{!}{
    \begin{tabular}{rrrr}
    \toprule[1pt]
    \multicolumn{1}{c}{Ques-Type} & \multicolumn{1}{c}{NE-based} & \multicolumn{1}{c}{KG-based} & \multicolumn{1}{c}{KG-NE} \bigstrut[b]\\
    \hline
    \multicolumn{1}{c}{Who} & \multicolumn{1}{c}{47975} & \multicolumn{1}{c}{94} & \multicolumn{1}{c}{24243} \bigstrut[t]\\
    \multicolumn{1}{c}{Where} & \multicolumn{1}{c}{19366} & \multicolumn{1}{c}{192} & \multicolumn{1}{c}{24352}  \\
    \multicolumn{1}{c}{What} & \multicolumn{1}{c}{4512} & \multicolumn{1}{c}{5} & \multicolumn{1}{c}{4510}  \\
    \multicolumn{1}{c}{When} & \multicolumn{1}{c}{15736} & \multicolumn{1}{c}{4033} & \multicolumn{1}{c}{24132}  \\
    \multicolumn{1}{c}{How} & \multicolumn{1}{c}{13911} & \multicolumn{1}{c}{39176} & \multicolumn{1}{c}{24263}  \bigstrut[b]\\
    \hline
    \hline
    \multicolumn{1}{c}{\textbf{Total}} & \multicolumn{1}{c}{101500} & \multicolumn{1}{c}{43500} & \multicolumn{1}{c}{101500} \bigstrut[t]\\
    \toprule[1pt]
    \end{tabular}%
    }
    \caption{The number of the generated questions, using NE-based/KG-based/KG-NE method, in terms of different types. }
  \label{tab:analysis2}%
\end{table}%
\vspace{-0.8cm}

\subsection{Quality Analysis of Synthetic Data}
We conducted a meticulous manual review of two fundamental aspects of the synthetic question: the quality of the generated questions and the appropriateness of the candidate options. Our evaluation revealed that a significant portion of the questions suffered from grammatical errors and lacked contextual alignment. Moreover, the NE-based method often produced candidate options that did not align well with the questions, although there were exceptions. In contrast, the KG-based method consistently maintained semantic coherence between questions and candidate options. However, it primarily generated `How-type' questions and exhibited an overall lower question quantity. Finally, our findings suggest that the KG-NE method outperforms alternative methods in our study, offering a more promising approach.

\section{CONCLUSION}

In this work, we handle the MCQA task in an unsupervised manner under a fully non-annotated scenario, where no target data is given and only universal corpus can be utilized.
We propose a two-stage framework featured with question-answer pair generation and KG-NE based distractor generation, to construct the synthetic data for model training. 
The experimental results on multiple datasets verity the effectiveness of our approach and the impact of various distractor generation methods.

\vfill\pagebreak

\small
\bibliographystyle{IEEEbib}
\bibliography{Template}

\end{document}